\documentclass[lettersize,journal]{IEEEtran}
\usepackage{amsmath,amsfonts}
\usepackage{algorithmic}
\usepackage{algorithm}
\usepackage{array}
\usepackage[caption=false,font=normalsize,labelfont=sf,textfont=sf]{subfig}
\usepackage{textcomp}
\usepackage{stfloats}
\usepackage{url}
\usepackage{verbatim}
\usepackage{graphicx}
\usepackage{cite}

\usepackage{xcolor}

\usepackage{color}
\usepackage{amssymb}
\usepackage{multirow}

\begin{document}

\title{The ProfessionAl Go annotation datasEt (PAGE)}

\author{Yifan Gao, Danni Zhang, Haoyue Li
\thanks{Yifan Gao is the corresponding author (e-mail: yifangao@mail.ustc.edu.cn).}
\thanks{Yifan Gao is with School of Biomedical Engineering, University of Science and Technology of China, Hefei, China. Danni Zhang is with School of Economics, Anhui University, Hefei, China. Haoyue Li  is with College of Medicine and Biological Information Engineering, Northeastern University, Shenyang, China.}
}

\markboth{Journal of \LaTeX\ Class Files,~Vol.~14, No.~8, August~2021}%
{Shell \MakeLowercase{\textit{et al.}}: A Sample Article Using IEEEtran.cls for IEEE Journals}


\maketitle

\begin{abstract}
The game of Go has been highly under-researched due to the lack of game records and analysis tools. In recent years, the increasing number of professional competitions and the advent of AlphaZero-based algorithms provide an excellent opportunity for analyzing human Go games on a large scale. In this paper, we present the ProfessionAl Go annotation datasEt (PAGE), containing 98,525 games played by 2,007 professional players and spans over 70 years. The dataset includes rich AI analysis results for each move. Moreover, PAGE provides detailed metadata for every player and game after manual cleaning and labeling. Beyond the preliminary analysis of the dataset, we provide sample tasks that benefit from our dataset to demonstrate the potential application of PAGE in multiple research directions. To the best of our knowledge, PAGE is the first dataset with extensive annotation in the game of Go. This work is an extended version of \cite{cog} where we perform a more detailed description, analysis, and application.
\end{abstract}


\section{Introduction}
Go (weiqi, baduk) is one of Asia's most popular board games, especially in China, Japan, and Korea \cite{togo}. In the last two decades, professional Go tournaments have grown dramatically, with millions of audiences watching games from TV streams or online servers.

Data-driven analytics in traditional sports, e-sports, and board games has been a popular research area \cite{happennext,premierleague,combining,tabletennis,moba}. These evolving analytics techniques have significantly contributed to the sport and game community. In recent years, with the development of game record databases, large-scale analysis of Go has become a reality. In particular, the advent of AlphaZero \cite{alphazero} and KataGo \cite{katago} has made it possible to understand human decision-making in games at a deep level.

In such background, the analysis of Go is potentially very promising. On the one hand, as a popular competitive game, developing data-driven analytic technologies for Go can enrich the fan experience, help players improve their abilities, and promote other favorable aspects. On the other hand, as a game with simple rules and complex content, it is a valuable vehicle for psychological research \cite{lifego,humango}. However, conducting such research on Go remains nontrivial and challenging, and there is an extreme lack of relevant work. Firstly, there are no structured professional datasets. It is difficult for researchers to extract metadata and other game information. Secondly, no statistics on Go matches can directly demonstrate the state of both sides (e.g., shots in soccer). Finally, few people know about Go compared to popular sports and games, so it is not easy to organize data and construct practical features well.

To overcome these problems, we present the ProfessionAl Go annotation datasEt (PAGE), containing 98,525 games played by 2,007 professional players from 1950 to 2021. The raw records and metadata of the games are derived from the publicly available Go database. We annotate the metadata related to players and tournaments by combining multiple trustworthy sources.  A comprehensive in-game statistics feature is calculated by KataGo, which analyzes all games. As a result, the dataset is of good quality and has broad coverage.

In this paper, we provide three challenging downstream tasks to demonstrate the application of PAGE in several research directions. First, we used PAGE to evaluate the relationship between gender differences in ratings and participation rates of professional Go players. Second, several Convolutional Neural Networks (CNNs) and Transformer architectures were applied to predict blunders of professional players. Finally, we used several popular machine learning methods to predict game outcomes from historical data. The experimental results show that PAGE can potentially be applied in different studies.

In summary, the main contributions of our work are:
\begin{itemize}
	\item We present the first professional Go dataset with extensive annotation. The dataset contains a large amount of metadata and in-game statistics, facilitating massive research. The dataset will be made publicly available\footnote{\url{https://github.com/YifanGao00/The-Professional-Go-Dataset}}. 
	
	\item We describe several applications that benefit from our dataset and suggest possible research directions.
\end{itemize}

This work is an extended version of \cite{cog}, where additional explanations, comparisons, and applications are provided. We add extra in-game statistics features, including uncertainty and ownership, compared to the previous work. In addition, this work explains in more detail the process of building PAGE, as well as the introduction of the dataset structure. In particular, we also add two applications, containing the analysis of participation rates in gender differences and the blunder prediction of professional players.

The rest of the paper is organized as follows: Section II presents the background related to the proposed dataset. In section III, we describe the annotation procedure of PAGE in detail and provide the preliminary analysis of the dataset. Section IV presents three applications of the analysis performed on our dataset: participation analysis, blunder prediction, and game outcome prediction. Section V discusses potential research directions and limitations of PAGE, and the paper concludes in Section VI.

\begin{figure*}[!t]
	\centering
	\includegraphics[width=7in]{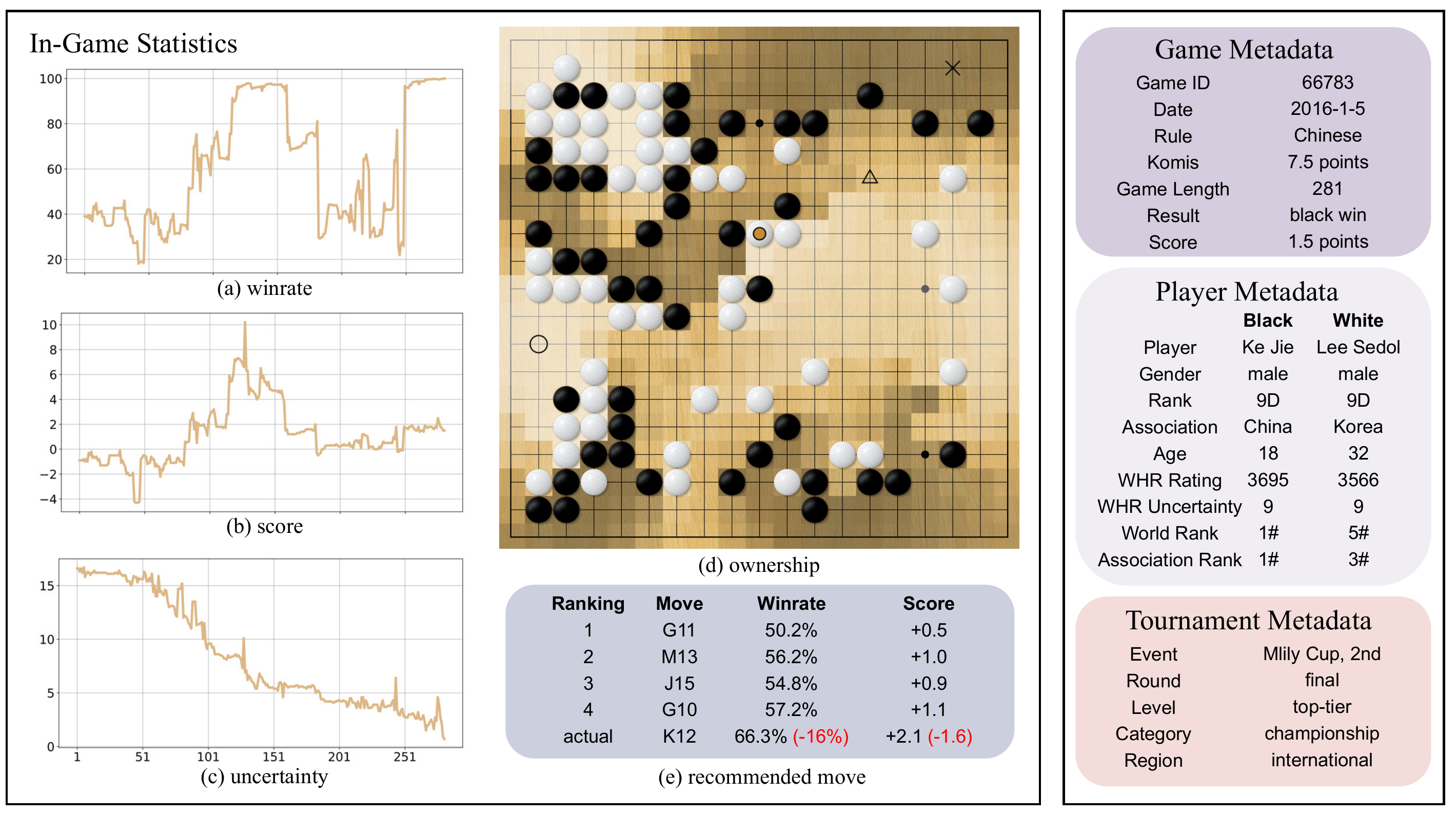}
	\caption{In-game statistics and metadata in the game between Ke Jie and Lee Sedol in the final of the 2nd Mlily Cup. It is the core property of the PAGE.}
	\label{fig1}
\end{figure*}

\section{Background}

\subsection{Professional Go}
Even though Go originated in China, the professional Go system made its appearance in Japan for the first time. Go became a popular game in Japan during the Edo period since the Shogun funded its development. In the early 20th century, Japan became the first country to establish a modern professional system, maintaining its absolute leadership until the 1980s. Nie Weiping won the Japan-China Super Go Tournament in 1985 after an incredible streak of victories against top Japanese players. Four years later, a highly anticipated Ing Cup final saw Cho Hunhyun defeat Nie Weiping 3:2 and win the 400,000 dollar prize. As a result of these two landmark events, China, Japan, and Korea formed a triple balance of power. In 1996, Korean Go entered its golden age, which lasted for ten years. Korean players, represented by Lee Changho, won most of the tournament titles during this decade. Chinese players progressed rapidly afterward and became fierce competitors with Korean players.

A couple of the earliest Go world tournaments were the Ing Cup and the Fujitsu Cup, which were founded in 1988. The Go world champions are the highest honor for professional players, similar to the Grand Slam in tennis. In addition to this, there are a large number of regional-level competitions, mainly consisting of single-elimination tournaments and leagues.

\subsection{Computer Analysis of Human Games}
With AlphaZero outperforming the best professional players by a large margin, researchers are focusing on more challenging tasks. However, AlphaZero and its open-source implementations, like ELF OpenGo \cite{elfopengo} and Leela Zero \cite{leelazero}, cannot be used to analyze human games precisely. According to their evaluation, they only take into account the win rate. Their win rate fluctuates dramatically when there is a small score gap, so the robustness is not good. Furthermore, these AIs cannot adjust for different rules and komis (points added to compensate for the black player's first move advantage), resulting in incorrect game evaluations.

KataGo is the state-of-the-art AlphaZero-based framework. In addition to providing rich in-game statistics, it uses improved technologies for more precise analysis. It supports various rules and komis. Consequently, KataGo is appropriate for analyzing human games and obtaining in-game statistics. KataGo supports many different statistics, but PAGE focuses primarily on five: win rate, score difference, uncertainty, ownership, and recommended moves.

Fig. \ref{fig1}a to Fig. \ref{fig1}e shows the schematic diagram of the in-game statistics. In this case, uncertainty and ownership are fine-grained features unique to KataGo. In particular, the uncertainty is the variance of the score in the current board state. Due to the characteristics of Monte Carlo Tree Search (MCTS), this value is significantly large but still serves as a good indicator of uncertainty. Ownership is the expected ownership of each position on the board, where 1 denotes the current player's ownership and -1 denotes the opponent's ownership. As can be seen in Fig. \ref{fig1}d, the closer to black, such as multiplication symbol in the figure, is to the area occupied by black. Conversely, the closer to white, such as the area of the circle, is more likely to be white's territory. The triangular areas in the diagram represent the areas that both players need to fight over.

\subsection{Player Performance Analysis}
It has been a long-standing practice for professional Go players to improve themselves and prepare for tournaments using intuition-driven performance analysis methods. Players will review their opponent's matches, recognize their weaknesses, and expect to gain an advantage in upcoming games by replaying and reviewing their games. Recently, AlphaZero and its modified algorithms have become far superior to humans in Go. Its incredible insight into the game and accurate evaluation capabilities are helping many professional Go players improve their performance. Despite this, players rarely research further based on purposefully collected data, so these analysis methods are still intuition-driven.

There are well-established and promising data-driven analytics methods for many sports and games. Merhej~\cite{happennext} uses deep learning to assess the value of defensive behavior in football. Baboota~\cite{premierleague} predicts Premier League outcomes through machine learning and gets good results. Beal~\cite{combining} presents a novel natural language processing technology, combining the statistics of sports reporters with background articles to improve predictive performance at football games. In Castellar's study \cite{tabletennis}, reaction time and exercise time are examined as factors affecting the performance of table tennis players. Yang~\cite{moba} proposes an explainable two-stage network for predicting real-time win rates in multiplayer online battle arena games. These analysis techniques are vital to improving the fan experience and assessing players' performance.

The AlphaZero algorithm is used in many board games, including Gomoku \cite{gomokunet}, Othello \cite{othello}, and NoGo \cite{nogo}. However, these games do not have many accessible game records, making it challenging to develop analysis techniques for them. In contrast, the detailed chess datasets, many participants, and advanced chess computer engines significantly impacted chess analytics. Skill evaluations \cite{skillperformance,skillcomputer}, ratings \cite{intrinsic,actualplay,evolutionaryrating}, and style modeling \cite{preference,mcilroy2020aligning,mcilroy2020learning,mcilroy2021detecting} are the most common applications. These technologies demonstrate the great potential of player performance analysis in the board game. There is little previous research on the game of Go. Over the last two decades, however, a growing number of tournaments, players, and recorded games has allowed the development of a dataset and benchmarks for Go. As a result, we propose PAGE, a large-scale professional Go annotation dataset. Using this dataset, we hope to bridge the gap between this unique game and the game community.

\subsection{Board Game in Psychological Research}
Board games are an essential research part of psychological research, especially cognitive science. In particular, chess is a valuable tool for psychological investigations \cite{blanch2015sex}. Unlike other board games, chess has a long history of professionalization and a large number of well-organized and structured game records spanning decades. Researchers interested in various psychological topics can use these databases to conduct their research. In the following content, we present the application of the chess database in two topics, including the role played by gender and age in cognitive performance.

Howard~\cite{howard2006complete} highlighted the methodological value of chess data for the first time and presented a database with millions of records providing performance data on over 60,000 players since 1970. This article also looked at other intellectual games, such as bridge, Go, and backgammon. However, data for these games tend to be more limited than for chess, making it difficult to use as a tool for psychological research. By analyzing the chess database, Howard~\cite{howard2005gender} proposed that gender differences in elite mind sports arise from differences in innate ability. Bilalic~\cite{bilalic2006intellectual} countered this idea and argued that different participation rates or differences in the amount of practice, motivation, and interest of male and female players might better explain the gender differences in chess performance. Following this, more work focused on the relationship between participation rates and gender differences \cite{participation1,participation2,participation3}. Blanch~\cite{blanch2015sex} used the chess database to comprehensively analyze gender differences in chess from multiple perspectives. The findings demonstrate that biosocial factors, such as age and practice, rather than differences in participation rates, influence gender differences in Elo ratings. In addition to intelligence and participation rates, another reasonable explanation for the gender difference is the gender stereotype threat. Stafford~\cite{stafford2018female} analyzed a database of over 5.5 million chess matches and found no evidence of the stereotype threat effect. On the other hand, Smerdon~\cite{smerdon2020female} demonstrated that a typical gender stereotype threat exists in chess by introducing the variable of the opponent's age. Furthermore, some studies have used large-scale chess databases to verify psychological effects, such as the gender equality paradox \cite{vishkin2022queen}.

Chess is also often used as a tool to study the effects of age on cognitive performance. Fair~\cite{fair2007estimated} estimated decline rates for chess and various sports event and confirmed that the decline was much smaller for chess than for sports. Roring~\cite{roring2007multilevel} used a multilevel modeling approach to a large database of chess players to study longitudinal changes in chess skills. Vaci~\cite{vaci2015age} used linear mixed effects models to explore the hypothesis that "age is more friendly to the more able." In the latest study, Strittmatter~\cite{strittmatter2020life} presented evidence for a life-cycle model of cognitive performance based on a comprehensive analysis of professional chess tournament data.

Large-scale chess databases have played a crucial role in these studies. However, due to the limitations of the dataset, most of the information used in these studies contained only ELO ratings \cite{elo} and game results. With the ongoing development of AlphaZero-based algorithms, AI has been able to provide more insight into human decision-making within the game. Specifically, comprehensive and detailed fine-grained statistics, such as win rates, score differences, uncertainty, and ownership contained in PAGE, have the potential to provide more convincing evidence for psychological studies.

\begin{figure*}[!t]
	\centering
	\includegraphics[width=7in]{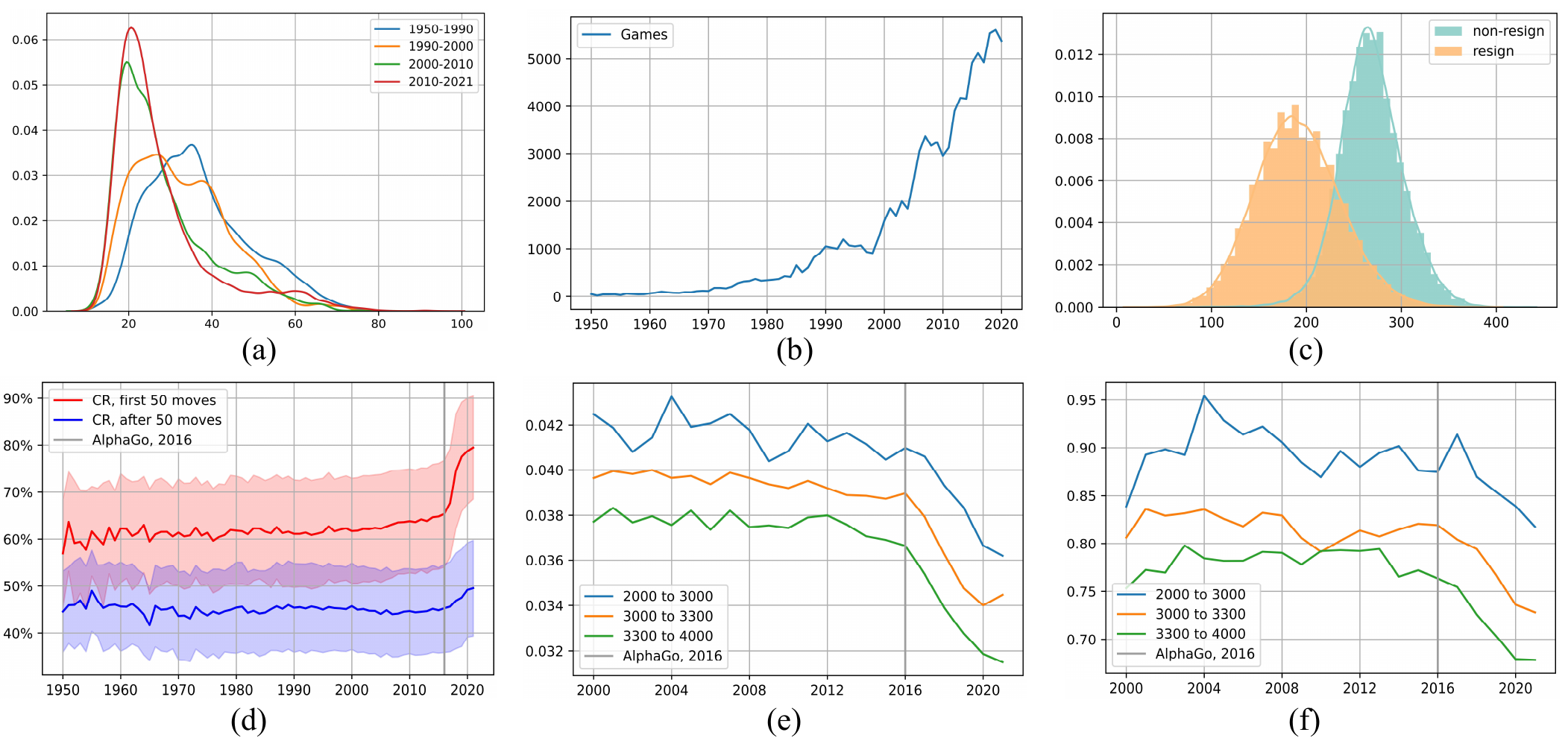}
	\caption{Illustrations of dataset statistics. (a) distribution of the age of players in different generations; (b) game counts in years; (c) distribution of game lengths; (d) Mean move similarity in years; (e) mean loss win rate with different ratings in years; (f) mean loss score with different ratings in years.}
	\label{fig2}
\end{figure*}

\section{The Professional Go Annotation Dataset}

\subsection{Data Acquisition}
\subsubsection{Metadata}
We begin with the raw game data by accessing Go4Go\footnote{\url{https://www.go4go.net}}, the largest Go game database. Game records in the database can be used for academic research with the owner's permission. Some games were excluded from the raw records, including AI competitions, amateur games, handicap games, and games that ended abnormally. In the end, we obtained 98,525 games played by 2,007 professional players.  We extracted metadata about each player’s birth date, gender, and affiliated association from three reliable online references: Go Ratings\footnote{\url{https://www.goratings.org/en/}}, Sensei’s Library\footnote{\url{https://senseis.xmp.net}}, and List of Go Players\footnote{\url{https://db.u-go.net}}

Metadata related to matches and rounds were extracted from SGF files. We finally got 503 tournament categories (e.g., Chunlan Cup, LG Cup), 3131 tournaments (e.g., 1st Chunlan Cup, 15th LG Cup), and 342 rounds (e.g., round 2, final). We extracted data regarding matches and rounds from SGF files. Ultimately, we found various tournament categories (e.g., Chunlan Cup, LG Cup), tournaments, and rounds (e.g., round 2, final). Every tournament was manually labeled by a knowledgeable amateur player, including the category, level, and region. The tournament categories include championship, league, team, and friendly. The level of tournaments includes top-tier and regular. The region includes international and regular.

We used a Python script\footnote{\url{https://github.com/pfmonville/whole_history_rating}} to calculate the player's WHR ratings \cite{whr}. To obtain a fine-grained rating score, we took the following calculation approach. First, the initial number of iterations was set to 50, Second, by adding games day by day and performing one iteration, the rating was calculated for all time points. Notably, following the Go Ratings, the initial score was set to 3000. Finally, we obtained the WHR rating and the uncertainty. The uncertainty is higher if the player has just entered the leaderboard or has not played for too long.

\subsubsection{In-game Statistics}
The game records were analyzed using KataGo v1.9.1 with the TensorRT backend. In order to achieve a balance between accuracy and speed, the following strategy was applied: We performed 100 simulations for each move to obtain initial in-game statistics. After a move with a high fluctuation (greater than 10\% win rate or 5 points), KataGo would reevaluate it with a simulation count of 1000. Finally, we obtained in-game statistics for all games, including win rate, score difference, uncertainty, ownership distribution, and recommended moves. We conducted our analysis on an NVIDIA RTX 2080Ti graphic card, which took about 40 days.

\begin{table}
	\centering
	\renewcommand\arraystretch{1.2}
	\caption{Most frequent players.}
	\begin{tabular}{cc|cc}
		\hline
		Players           & Games & Players           & Games  \\ \hline
		Cho Chikun       & 2079       & O Rissei         & 1217        \\
		Lee Changho      & 1962       & Yamashita Keigo  & 1217        \\
		Kobayashi Koichi & 1605       & Gu Li            & 1215       \\
		Rin Kaiho        & 1536       & Otake Hideo      & 1214       \\
		Cho Hunhyun      & 1533       & Cho U            & 1145       \\
		Lee Sedol        & 1375       & Choi Cheolhan    & 1133       \\
		Yoda Norimoto    & 1316       & Chang Hao        & 1100       \\
		Kato Masao       & 1249       & Park Junghwan    & 1058       \\
		Takemiya Masaki  & 1248       & Kobayashi Satoru & 1044       \\ \hline
	\end{tabular}
	\label{tab:frequentplayers}
\end{table}

\begin{table}
	\centering
	\renewcommand\arraystretch{1.2}
	\caption{Most frequent matchups.}
	\begin{tabular}{cccc}
		\hline
		Matchups                        & Games & Win & Loss \\ \hline
		Cho Hunhyun vs. Lee Changho     & 287   & 110 & 177  \\
		Cho Hunhyun vs. Seo Bongsoo     & 207   & 139 & 68   \\
		Kobayashi Koichi vs. Cho Chikun & 123   & 60  & 63   \\
		Yoo Changhyuk vs. Lee Changho   & 122   & 39  & 83   \\
		Cho Chikun vs. Kato Masao       & 107   & 67  & 40   \\ \hline
	\end{tabular}
	\label{tab:frequentmatchups}
\end{table}

\begin{table}
	\centering
	\renewcommand\arraystretch{1.2}
	\caption{Most frequent tournaments.}
	\begin{tabular}{cc|cc}
		\hline
		Tournaments       & Games & Tournament             & Games \\ \hline
		Chinese League A & 11092 & Japanese Oza           & 1996  \\
		Korean League A  & 4773  & Japanese NHK Cup       & 1939  \\
		Japanese Honinbo & 3655  & Samsung Cup            & 1806  \\
		Japanese Ryusei  & 2932  & Chinese Mingren        & 1588  \\
		Japanese Meijin  & 2869  & Chinese Women's League & 1569  \\
		Japanese Judan   & 2766  & LG Cup                 & 1514  \\
		Japanese Kisei   & 2723  & Korean League B        & 1411  \\
		Japanese Tengen  & 2383  & Chinese Tianyuan       & 1282  \\
		Japanese Gosei   & 2230  & Korean Women's League  & 1279  \\ \hline
	\end{tabular}
	\label{tab:tournaments}
\end{table}

\subsection{Dataset Analysis}
In this section, we introduce the preliminary analysis of PAGE. There are two aspects to the statistics: players and games.
\subsubsection{Players}
Fig. \ref{fig2}a presents the age distribution of the players in the different generations of the games. The average age of players before 1990 was between 30 and 50. As the decade progressed, the average age of players dropped rapidly to less than 30. The majority of competitions were completed by players in their 20s in the 21st century. It is evident from this trend that professional Go is becoming more competitive. Table \ref{tab:frequentplayers} and Table \ref{tab:frequentmatchups} show the most frequent players and matchups. Most of the games were completed by these prominent legends, illustrating a long-tailed distribution of games played by different players.
\subsubsection{Games}
Fig. \ref{fig2}b shows the number of games per year in the dataset. As we can see from the graph, it is growing exponentially over time. On the one hand, it indicates that the number of professional matches has increased in the last couple of years. On the other hand, PAGE does not appear to have recorded many games during the early years. Table V shows the tournaments with the greatest number of games played, mostly Japanese tournaments, except for the Chinese League A and the Korean League A. The main reason is that these tournaments have been held for a long time. As shown in Fig. \ref{fig2}c, resigned games have shorter lengths than non-resigned games. Fig. \ref{fig2}d to \ref{fig2}f show some trends of in-game statistics over time. We observe some interesting results from these schematics. With AlphaGo's advent, professional players began to imitate the AI's preferred moves, which is particularly evident in the opening phase (first 50 moves). One can observe the coincidence rate in Fig. \ref{fig2}d, which grew significantly after 2016. In addition, the non-opening phase is difficult to imitate, so the coincidence rate grows much slower than in the opening stage. According to Fig. \ref{fig2}e and Fig. \ref{fig2}f, the average loss of win rate and the average loss of score also decreased after AlphaGo emerged. Furthermore, players with higher WHR ratings have lower statistics than those with lower scores. These characteristics offer the potential to develop player performance analysis techniques.

\section{Downstream Tasks}

\subsection{Participation Analysis}
In this section, we use PAGE to evaluate the relationship between gender differences and participation rates in professional Go. In chess, recent psychological research has proven that gender differences are not related to differences in participation rates. First, we describe the methodology used and the details of the data. Second, we report the experimental results.

\begin{figure}[!t]
	\centering
	\includegraphics[width=3.5in]{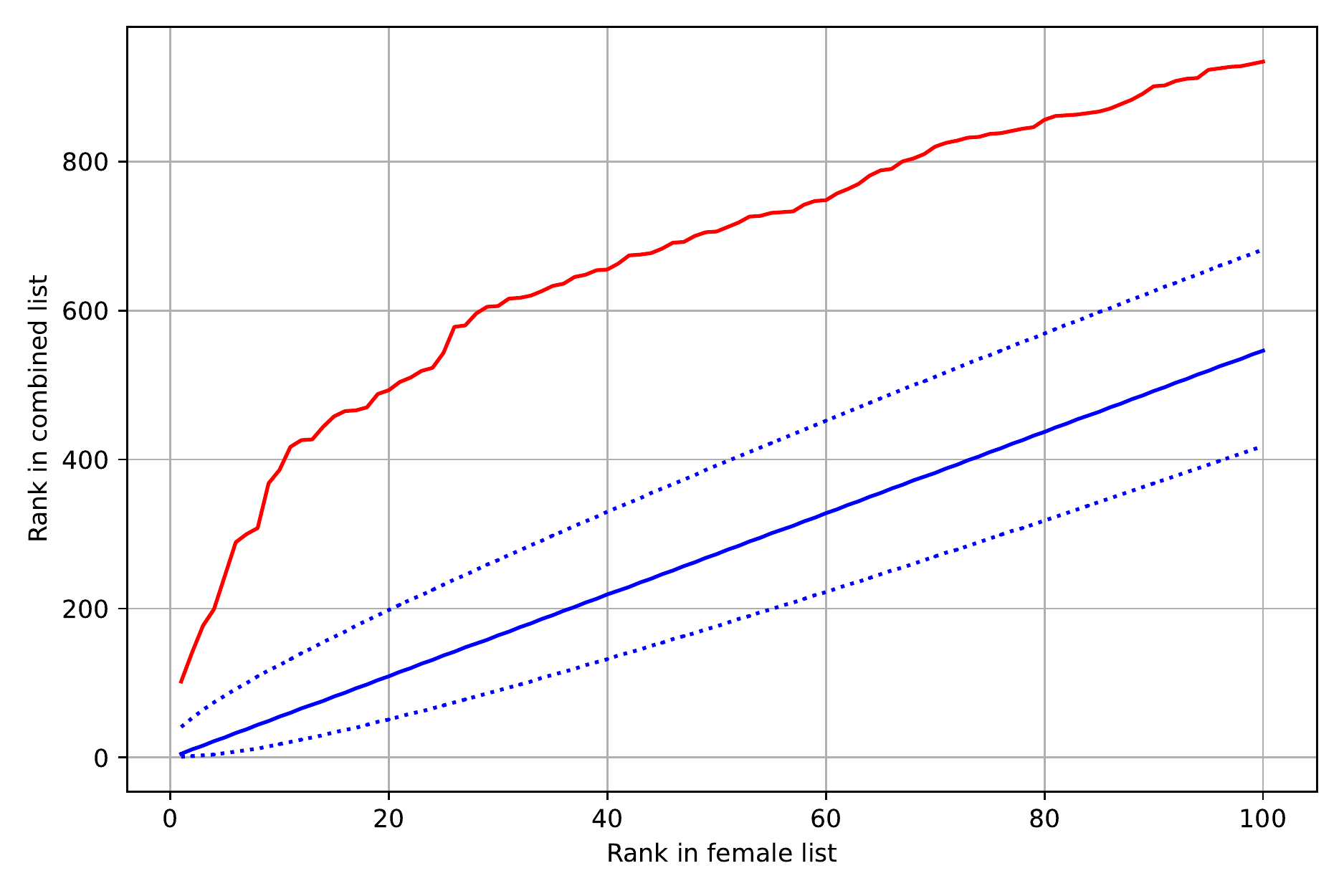}
	\caption{Observed and expected rankings of the WHR ratings for the top 100 female players. The Red line is the exact rank, and the blue line is the expected rank. The dotted lines represent the quantiles $r_{low}$ and $r_{high}$.}
	\label{fig3}
\end{figure}

\begin{figure}[!t]
	\centering
	\includegraphics[width=3.5in]{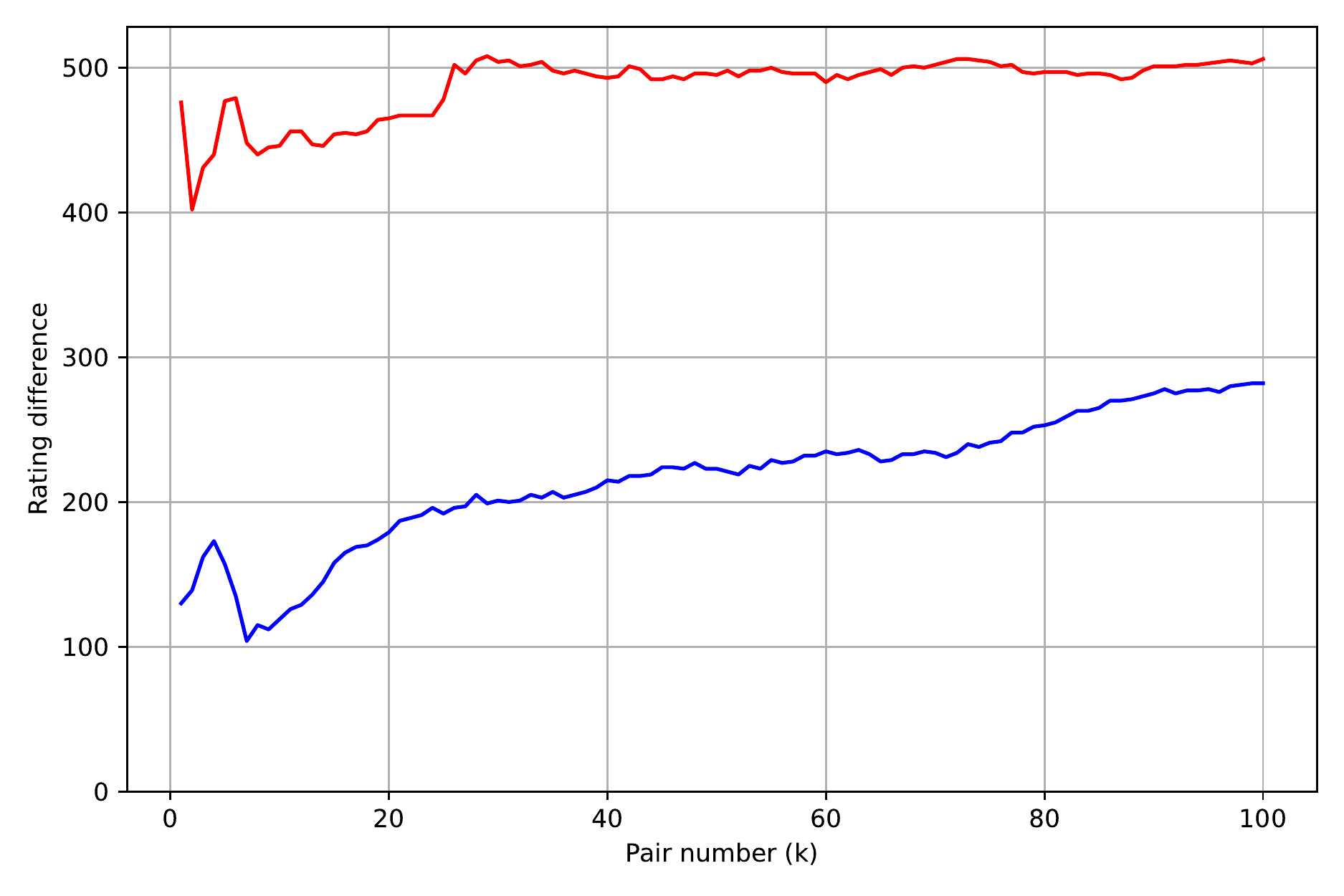}
	\caption{Observed and expected sex differences in WHR ratings. The red line is the actual score differences, and the blue line represents score differences attributed to different participation rates.}
	\label{fig4}
\end{figure}

\begin{figure}[!t]
	\centering
	\includegraphics[width=3.5in]{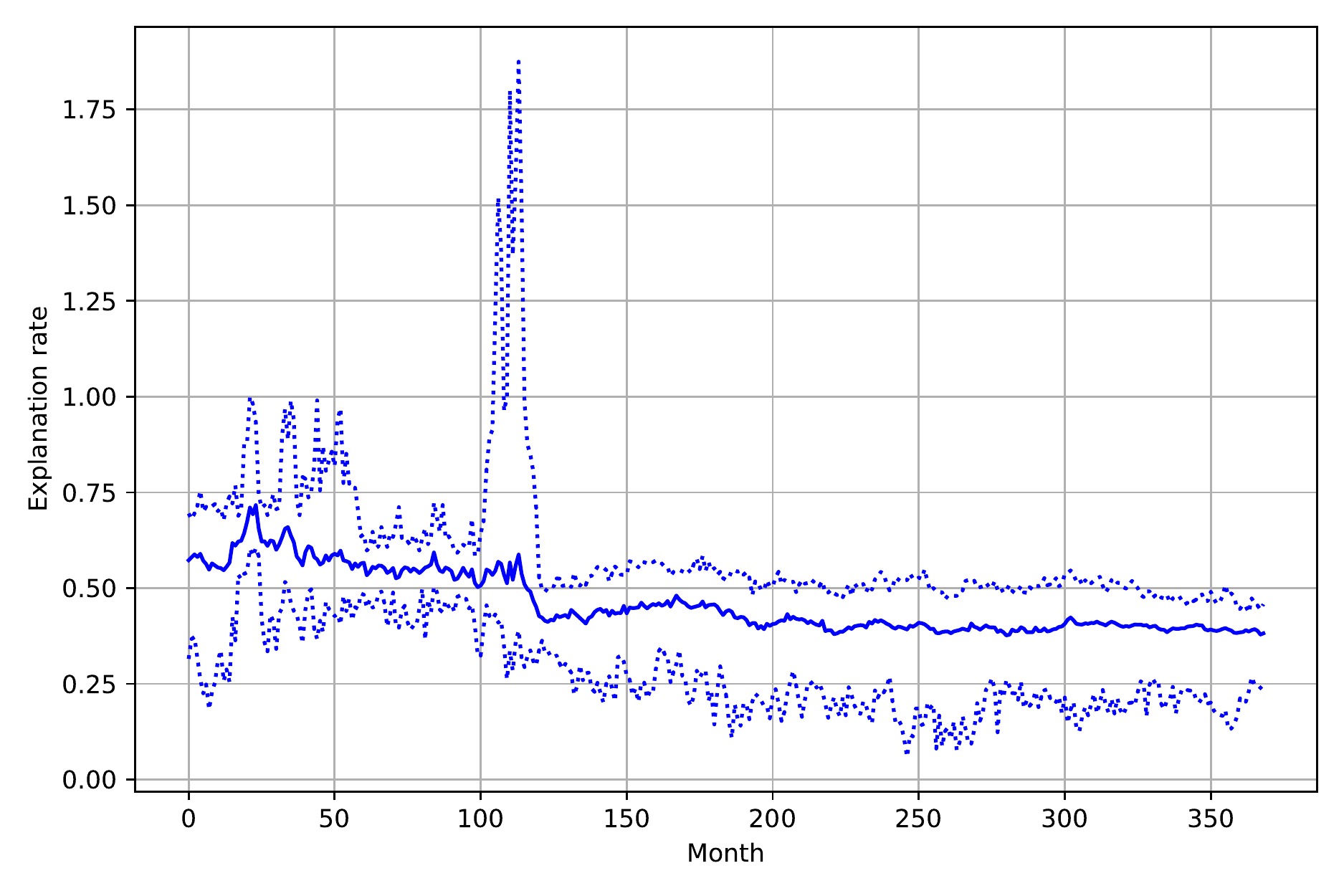}
	\caption{The month-by-month trend of the explanation rate of the actual rating differences. The dotted lines represent the upper and lower bounds of the observed values.}
	\label{fig5}
\end{figure}

\subsubsection{Materials and Methods}
Inspired by the work of Knapp \cite{participation3} and Blanch \cite{blanch2015sex}, we conducted three experiments to explore the relationship between differences in participation rates and gender differences. First, we tested WHR scores using the negative hypergeometric distribution. Specifically, among $N$ female and $M$ male players, $R_k$ was the best female player ranked $k$-th among all combinations of participants. We calculated the 0.05\% and 99.5\% quartiles of the expectation distribution $r_{low}$ and $r_{high}$. If there was no gender effect in Go performance, the ranking $R_k$ of the $k$-th ranked female player should be between $r_{low}$ and $r_{high}$. Second, we made a binary observation for each sampling unit to determine the weight of the difference in rating scores that can be attributed to the difference in participation rates between male and female players, i.e., the explanation rate. Finally, we performed month-by-month calculations of the explanation rate to investigate changes in the gender difference over time.

In the first and second experiments, we selected the WHR ratings for December 2021 and excluded players with less than ten games. Finally, 1366 players were included in the calculation, with 1116 male and 250 female players. The experiments examined the ranking of the top 100 best female players, i.e., $k$ was set to 100. In the third experiment, we calculated the explanation rate month-by-month, where at least 50 female players entered the rating that month. To reduce the effect of the rating distribution on the results, we selected only the top 20\% of female players each month and computed the average. For example, $k=10$ in April 1991 and $k=50$ in December 2021.

\subsubsection{Results}
Fig. \ref{fig3} shows the observed and expected rankings of the WHR scores for the top 100 female players. The observed rank scores (red line) lie outside the 99.5\% quantile of the expected ranking (blue line). The actual score differences (red line) and the score differences attributed to different participation rates (blue line) are shown in Fig. \ref{fig4}. The unexplained differences between the actual and expected differences range from 221 to 346 points, with a mean of 272. In other words, only 23.2\% to 56.1\% (mean 44\%) of the actual scoring difference is explained by the different participation rates of male and female players. Fig. \ref{fig5} shows the month-by-month trend of the explanation rate of the actual rating differences. As can be seen, there is an overall decreasing trend in the explanation rate. It is worth noting that the maximum value of the explanation rate has been very high for several months. This is because Rui Naiwei reached second place in the world during this period, which is also the highest ranking ever for a female player.

From the three experiments mentioned above, we can find that gender differences do exist in professional Go tournaments, and participation rates barely explain the differences in WHR scores. Compared to chess, Go tournaments are a unique tool for studying East Asian cultures. However, there is little psychological research on the game of Go. We present the first analysis of the relationship between participation rates and gender differences through PAGE and expect to motivate more psychological work in the future.

\begin{table}[]
	\centering
	\renewcommand\arraystretch{1.2}
	\caption{Experiment results in blunder prediction.}
	\begin{tabular}{cccc}
		\hline
		Model     & Accuracy (\%)   & Precision 1\# (\%)  & Precision 2\# (\%)  \\ \hline
		ConvNet   & 69.64          & 73.67          & 65.58          \\
		ZeroNet   & \textbf{71.22} & \textbf{79.34} & 65.06          \\
		KataNet   & 71.08          & 76.18          & 66.36          \\
		Perceiver & 71.10          & 74.66          & \textbf{67.38} \\ \hline
	\end{tabular}
	\label{tab:blunder}
\end{table}

\begin{table*}[]
	\centering
	\renewcommand\arraystretch{1.2}
	\caption{Features description in game outcome prediction.}
	\begin{tabular}{cc}
		\hline
		\textbf{Features}                & \textbf{Description  }                                                                                  \\ \hline
		\textbf{Metadata Features}       &                                                                                                \\
		Basic Information       & Include the games time, age, gender, association.                                               \\
		Ranks                   & Ranking of Go players.                                                                          \\
		WHR Score               & WHR rating, which measures the level of the player.                                             \\
		WHR Uncertainty         & A measure of uncertainty in the WHR rating system.                                             \\
		Tournament Feature      & Features of the tournament.                                                                    \\ \hline
		\textbf{Contextual Features}     &                                                                                                \\
		Match Results           & The player’s recent performance in various competitions.                                       \\
		Match Results by Region & Performance against the opponent’s region.                                                     \\
		Matchup Results         & Past competitions against the opponent.                                                         \\
		Tournament Results      & Past performances at this tournament.                                                          \\
		Opponents Ranks         & Rank of opponents.                                                                              \\
		Opponents Ages          & Age of opponents.                                                                               \\
		Cross-region Counts     & The number of crossregional competitions.                                                      \\ \hline
		\textbf{In-game Features}        &                                                                                                \\
		Mean Win Rate           & Average win rate in recent games. \\
		Mean Score              & Average score difference.                                                                       \\
		Mean Loss Win Rate      & Average win rate lost.                                                                          \\
		Mean Loss Score         & Average score lost.                                                                             \\
		Advantage Rounds        & Number of rounds with a 5\% win rate or a 3 point advantage.                                    \\
		Strong Advantage Rounds & Number of rounds with a 10\% win rate or a 5 point advantage.                          \\
		Coincidence Rate        & The ratio of moves that match KataGo’s recommendation to total moves.                     \\ \hline
	\end{tabular}
	\label{tab:feature}
\end{table*}

\subsection{Blunder Prediction}
Predicting the upcoming mistakes of human players from the board state is a promising task. In professional Go tournaments, top players often steer their opponents to positions that are more likely to make mistakes and gain an advantage through their opponent's blunders. However, due to Go's abstract and complex characteristics, directly analyzing human decision-making is highly challenging. In recent years, the development of deep learning has made it possible to model the fine-grained behaviors and decisions of human players. In this section, we leverage various deep learning methods, including CNNs \cite{he2016deep} and Transformer architectures \cite{dosovitskiy2021an,dai2021transmed}, to predict blunders in professional Go tournaments.

\subsubsection{Materials and Methods}
This study explored the performance with several state-of-the-art deep learning models, including ConvNet, ZeroNet, KataNet, and Perceiver Transformer. Shao~\cite{shao2016move} designed CNNs to predict the moves of the board game, and it has achieved satisfactory performance in the RenjuNet database. This work uses regular CNNs, so we name it ConvNet. We call the network used in AlphaGo Zero as ZeroNet. Furthermore, the architecture used in KataGo is KataNet. There are no domain-specific improvements, such as ownership subhead or scoring distribution, included in our implementation of KataNet. Last but not least, the Perceiver \cite{jaegle2021perceiver} is a model built on the Transformer architecture. The model utilizes an asymmetric attention mechanism that extracts the input iteratively into a latent bottleneck, allowing it to scale to handle multimodal inputs. To ensure fairness, we set the CNNs to roughly the same depth and width, namely 6 successive blocks and 64 channels. In the Perceiver, we adjust the hyperparameters to maintain the same order of magnitude of parameter size as the CNNs.

To construct the blunder detection as a classification problem, we define a blunder move as an action with a 10\% drop in win rate or a score loss of 5 points. Meanwhile, normal moves were consistent with the AI's recommended moves. We extracted a sample of 982,323 blunder moves from PAGE. On the other hand, we randomly selected 1,138,135 samples from the normal moves to form the blunder prediction dataset. We divided the training, validation, and test sets according to the ratio of 7:1:2. All deep learning models were trained for ten epochs. The batch size was set to 128, the optimizer was Stochastic Gradient Descent (SGD), and the learning rate was 0.01. The experiments were conducted on an NVIDIA RTX 3060 GPU with 12G video memory. For each method, we report the accuracy and precision of each class in the test set.

\subsubsection{Results}
Table \ref{tab:blunder} reports the performance of the four methods in predicting blunders. Among them, three metrics indicate the overall accuracy, the precision of normal move classification, and the precision of blunder move classification. The performance of ConvNet is lower than the other three methods, probably because the shallower network architecture does not capture the information and features of the board state well. ZeroNet performs the best in terms of accuracy, while Perceiver Transformer has the best precision in blunder move classification. In our experiment, we only used the base features, i.e., the current state of the board. While multimodal analysis combining more in-game statistics has the potential to improve the performance of blunder prediction significantly.

\subsection{Game Outcome Prediction}
This section takes a closer look at how we can use historical games to predict the outcome of future games based on historical data. Following feature extraction and pre-processing, we apply popular machine learning methods, and finally, we report the prediction system's performance.

\subsubsection{Materials and Methods}
First, we performed feature extraction and categorize all features into three groups: metadata features, contextual features, and in-game features. The detailed meanings of these features are illustrated in Table \ref{tab:feature}.

XGBoost \cite{xgboost}, Random Forest (RF) \cite{randomforest}, LightGBM \cite{lightgbm}, and CatBoost \cite{catboost} were selected as the training methods for the outcome prediction models. Three rating-based models were used for comparison: ELO, WHR, and TrueSkill \cite{trueskill}. In particular, we used the corresponding hyperparameters in the Python package by default and did not tune them. Although tuning these hyperparameters could improve the performance, our proposed approach has produced promising performance compared to other rating-based models.

\begin{table*}[]
	\centering
	\renewcommand\arraystretch{1.2}
	\caption{Experimental results in game outcome prediction. The best performance is indicated by bolded fonts. As can be seen in the red font, our proposed approach is significantly better than the best rating-based approach.}
	\begin{tabular}{lllllllllllll}
		\hline
		\multicolumn{1}{c}{} & \multicolumn{2}{c}{Mean}                            & \multicolumn{2}{c}{CR}                              & \multicolumn{2}{c}{CHN}                             & \multicolumn{2}{c}{KOR}                             & \multicolumn{2}{c}{JPN}                             & \multicolumn{2}{c}{Others}                          \\ \hline
		\multicolumn{1}{c}{} & \multicolumn{1}{c}{ACC↑} & \multicolumn{1}{c}{MSE↓} & \multicolumn{1}{c}{ACC↑} & \multicolumn{1}{c}{MSE↓} & \multicolumn{1}{c}{ACC↑} & \multicolumn{1}{c}{MSE↓} & \multicolumn{1}{c}{ACC↑} & \multicolumn{1}{c}{MSE↓} & \multicolumn{1}{c}{ACC↑} & \multicolumn{1}{c}{MSE↓} & \multicolumn{1}{c}{ACC↑} & \multicolumn{1}{c}{MSE↓} \\ \hline
		ELO                  & 0.6515                   & 0.2144                   & 0.6466                   & 0.2174                   & 0.6176                   & 0.2273                   & 0.6339                   & 0.2191                   & 0.6637                   & 0.2107                   & 0.7156                   & 0.1907                   \\
		WHR                  & 0.6567                   & 0.2125                   & 0.6684                   & 0.2090                   & 0.6212                   & 0.2295                   & 0.6397                   & 0.2193                   & 0.6577                   & 0.2108                   & 0.7254                   & 0.1813                   \\
		TrueSkill            & 0.6439                   & 0.2605                   & 0.6380                   & 0.2781                   & 0.6095                   & 0.1875                   & 0.6265                   & 0.2654                   & 0.6552                   & 0.2546                   & 0.7106                   & 0.2062                   \\ \hline
		XGBoost              & 0.7351                   & 0.1700                   & 0.7457                   & 0.1663                   & 0.7033                   & 0.1844                   & 0.7197                   & 0.1779                   & 0.7563                   & 0.1607                   & 0.7692                   & 0.1520                   \\
		RF                   & 0.6932                   & 0.2022                   & 0.6954                   & 0.1998                   & 0.6623                   & 0.2146                   & 0.6800                   & 0.2086                   & 0.7031                   & 0.1956                   & 0.7446                   & 0.1847                   \\
		LightGBM             & 0.7509                   & 0.1637                   & 0.7611                   & 0.1574                   & 0.7241                   & 0.1765                   & 0.7374                   & 0.1715                   & 0.7599                   & 0.1600                   & 0.7912                   & 0.1432                   \\
		CatBoost             & \textbf{0.7530}          & \textbf{0.1623}          & \textbf{0.7632}          & \textbf{0.1572}          & \textbf{0.7258}          & \textbf{0.1752}          & \textbf{0.7379}          & \textbf{0.1699}          & \textbf{0.7633}          & \textbf{0.1577}          & \textbf{0.7946}          & \textbf{0.1411}          \\
		& \textcolor{red}{9.6\%}           & \textcolor{red}{-0.050}          & \textcolor{red}{9.5\%}           & \textcolor{red}{-0.052}          & \textcolor{red}{10.5\%}          & \textcolor{red}{-0.052}          & \textcolor{red}{9.8\%}           & \textcolor{red}{-0.049}          & \textcolor{red}{10.0\%}          & \textcolor{red}{-0.053}          & \textcolor{red}{6.9\%}           & \textcolor{red}{-0.040}          \\ \hline
	\end{tabular}
	\label{tab:results}
\end{table*}

\begin{table}
	\centering
	\renewcommand\arraystretch{1.2}
	\caption{Ablation results in game outcome prediction.}	
	\begin{tabular}{ccccc}
		\hline
		Metadata & Contextual & In-game & ACC↑   & MSE↓   \\ \hline
		\checkmark        &           &        & 0.6719 & 0.1975 \\
		& \checkmark          &        & 0.7099 & 0.1827 \\
		&           & \checkmark       & 0.6883 & 0.1891 \\
		\checkmark        & \checkmark          &        & 0.7342 & 0.1706 \\
		\checkmark        & \checkmark          & \checkmark       & 0.7530  & 0.1623 \\ \hline
	\end{tabular}
	\label{tab:ablation}
\end{table}

\subsubsection{Results}
Table \ref{tab:results} shows the experimental results. We can see that WHR achieves an accuracy of 65.67\% and an MSE of 0.2125 among all rating system-based outcome prediction models, which is the best performer. Compared to ELO, WHR performs slightly better because it takes advantage of the long-term dependence on game results. There is a lower performance in TrueSkill because it is more suited to calculating ratings for multiplayer sports.

The best performance of the machine learning methods using various features is CatBoost, reaching an accuracy of 75.30\% and an MSE of 0.1623, which is much higher than WHR. The CatBoost model improves accuracy by 10\% and MSE by 0.05 in almost every category. The improvement of the Others category is lower because it is relatively easier to predict, while the WHR method has an accuracy of over 72\%.

We also conducted ablation experiments to check the validity of each modular feature using the CatBoost method. The results are shown in Table \ref{tab:ablation}. The model using only meta-information features is 67.19\%, which is slightly higher than the WHR method, suggesting that other attributes in meta-information, in addition to WHR ratings, play an important role in the prediction.

The accuracy was 70.99\% when only contextual features were used. In comparison, the accuracy of 68.83\% was achieved by using only in-game features, both higher than the state-of-the-art rating system-based prediction approach. By combining these features, the predictive ability of the model has improved. As demonstrated by the ablation experiments, the characteristics of each of our modules enhance the performance of the prediction system.

\section{Discussion}
Modeling, analyzing, and understanding human behavior and decision-making is nontrivial and challenging. In this paper, we present PAGE, containing fine-grained annotations of elite Go players for over seventy years. In the previous section, we applied the dataset to three downstream tasks and achieved satisfactory results. This section discusses more possible research directions and limitations of our proposed dataset.

\subsection{Possible research directions}
\subsubsection{Advanced in-game statistics}
In traditional sports, there are many advanced in-game statistics, such as expected goals (xG) in soccer and total points added (TPA) in basketball. These advanced statistics enhance the fan experience and allow for better analysis of player performance. Even though we can better analyze player performance using our extracted in-game features, we can still do better. The prediction ability could be enhanced by adding more advanced in-game features. For example, time series analysis of win rate, score, and uncertainty, can be improved with well-developed techniques.

\subsubsection{Behavior and style modeling}
In Go matches, fans pay attention most to the styles and behaviors of different players. Taking Lee Changho as an example, he was able to defeat his opponent at the last minute through extreme steadiness and control. Alternatively, Gu Li tries to fight his opponents every chance he gets.

Modeling player styles and understanding human decision-making are challenging but fascinating tasks. Several potential applications can be found through its use, including targeted training, preparation, and AI-assisted cheating detection \cite{cheating1,cheating2}. This problem has only been discussed in a relatively small amount of board game literature. In this article, Omori~\cite{shogi} proposes that shogi moves could be classified based on game style and that AI can be trained to match certain game styles. McIlroy-Young~\cite{mcilroy2020learning} presents a personalized model for predicting an individual's move and demonstrating how it captures human behavior at the individual level. The lack of proper analysis techniques and large datasets has contributed to the lack of noticeable success in style modeling and recognition. The advent of PAGE has helped to make progress on such issues.

\subsubsection{Rating system}
In a result prediction model, a player's winning percentage can only be predicted between two players. As opposed to this, a rating system can evaluate several players based on their relative strengths and is, therefore, more helpful in assessing a player's strengths. There are currently only two types of rating systems used for board games: win-loss and time. It is possible for board games with more features to evaluate a player's strength more effectively. Therefore, combining machine learning approaches with traditional rating systems has great potential.

\subsubsection{Live commentary enhancement}
Go matches are watched live on television or online platforms every year by millions of fans. A classic world Go tournament, on the other hand, has the tendency to last for over five hours in a room with just the host analyzing the game. This results in viewers getting bored with the show and stopping to watch. As AlphaZero has evolved, live Go broadcasts usually include the AI's estimation of the position, which improves the viewer experience. While showing the win rate is essential to draw viewers' attention, showing only the percentage of wins often leads to attacks on professional players because even top professionals' games are rated very negatively by AI. Our proposed PAGE can change this situation. In live streaming Go matches, PAGE has the potential to develop technologies like automatic commentary text generation, real-time statistics, and result prediction that will enable a more detailed and nuanced analysis, significantly improving the viewer experience in live streaming.

\subsection{Limitations}
Although our PAGE contains a wealth of statistics and metadata, there are still some weaknesses. First, due to computational resource limitations, we only provide in-game statistics for 100 simulations, which are unreliable in very few cases, especially in complex situations. Second, analysis using only KataGo may not always be reliable. In future work, we will look at longer computations and mutual validation of different AIs to improve the robustness of in-game statistics.

\section{Conclusion}
We present PAGE, the first professional Go dataset with extensive annotation. Our dataset provides a large number of valuable statistics and metadata that can be useful tools for various research topics, especially in game analytics and psychology investigations. The motivation for building this dataset is to facilitate and further promote research on a broad range of human-centered computing through this large, diverse, and accessible dataset. This work is an extended version of \cite{cog} in which we performed a more detailed dataset description and added two applications that benefit from PAGE.



\bibliographystyle{IEEEtran}
\bibliography{bare_jrnl_new_sample4}








\end{document}